%% file: SparseBO.tex
\documentclass{article} 
\usepackage{nips14submit_e,times}
\usepackage{hyperref}
\usepackage[sorting=none,style=numeric,backend=bibtex]{biblatex}
\usepackage{url}
\usepackage{verbatim}
\usepackage{graphicx}
\usepackage{caption}
\usepackage{subcaption}
\usepackage[style=long,smallcaps]{glossaries}
\usepackage{amsfonts}
\usepackage{amssymb}
\usepackage{mathtools}
\usepackage{xfrac}
\usepackage{booktabs}

\usepackage{framed}
\usepackage{xspace}
\newcommand{\note}[1]{}

    \DeclareMathOperator*{\argmax}{arg\,max}

    \newcommand{\acro}[1]{\textsc{\MakeLowercase{#1}}}

    \newcommand{\ya}{y_{+}}
    \newcommand{\xa}{x_{+}}

    \newcommand{\N}{\mathcal{N}}
    
    \newcommand{\D}{\mathcal{D}}
    \newcommand{\sX}{\mathcal{X}}

    \newcommand{\reals}{\mathbb{R}}
    
    \renewcommand{\O}{\mathcal{O}}

\renewcommand{\note}[1]{
    ~\\
    \frame{
        \begin{minipage}[c]{\textwidth}
            \vspace{2pt}\center{#1}\vspace{2pt}
        \end{minipage}
    }
    \\[3pt]
    }

\title{Automated Machine Learning on Big Data using Stochastic Algorithm Tuning}

\author{
Thomas Nickson, Michael A Osborne, Steven Reece and Stephen Roberts\\
MLRG\\
Department of Engineering Science\\
University of Oxford \\
\texttt{ \{tron,mosb,reece,sjrob\}@robots.ox.ac.uk} \\
}

 \nipsfinalcopy 
\input{glossary}

\begin{document}

\maketitle

\begin{abstract}

    We introduce a means of automating machine learning (\textsc{ml}) for big data tasks, by performing scalable stochastic Bayesian optimisation of \textsc{ml} algorithm parameters and hyper-parameters. More often than not, the critical tuning of \textsc{ml} algorithm parameters has relied on domain expertise from experts, along with laborious hand-tuning, brute search or lengthy sampling runs. Against this background, Bayesian optimisation is finding increasing use in automating parameter tuning, making \textsc{ml} algorithms accessible even to non-experts. However, the state of the art in Bayesian optimisation is incapable of scaling to the large number of evaluations of algorithm performance required to fit realistic models to complex, big data. We here describe a stochastic, sparse, Bayesian optimisation strategy to solve this problem, using many thousands of noisy evaluations of algorithm performance on subsets of data in order to effectively train algorithms for big data. We provide a comprehensive benchmarking of possible sparsification strategies for Bayesian  optimisation, concluding that a Nystr\"{o}m approximation offers the best scaling and performance for real tasks. Our proposed algorithm demonstrates substantial improvement over the state of the art in tuning the parameters of a Gaussian Process time series prediction task on real, big data.
\end{abstract}
\section{Introduction}

The performance of many machine learning algorithms is highly sensitive to the learning of model structure, parameters and hyperparameters \cite{hutter2014efficient}. Examples include parameters specifying model capacity (numbers of layers), learning rates and kernel parameters (bandwidths). Selecting these parameters has traditionally required deep domain expertise, brute-force search and/or laborious hand-tuning. As such, as the demand for machine learning algorithms grows faster than the supply of machine learning experts, there is an increasing need for methods to automatically configure algorithms to the task at hand, even where the task at hand is complex and algorithm performance (such as cross-validation error, or likelihood) expensive to evaluate. Ultimately, we aim to deliver off-the-shelf algorithms without compromising model quality: This is the challenge of \emph{automated machine learning}.

One approach to automated machine learning is the use of a global optimisation algorithm to automatically optimise the performance of a supplied algorithm as a function of its parameters \cite{bergstra2011algorithms}. One popular choice of global optimiser for this purpose is \gls{bo} \cite{brochu2010tutorial}, providing a robust means of exploring complex, multi-modal, performance surfaces. \gls{bo} has been employed successfully in the configuration of deep neural network hyperparameters \cite{snoek-etal-2012b,bergstra2011algorithms}, a complex vision architecture \cite{BerYamCox13}, and in Auto-\textsc{weka} \cite{ThoEtAl13}, a framework incorporating a diverse range of classification algorithms.
Most applications of \gls{bo} make use of a \gls{gp} surrogate for the objective function, enabling flexible non-parametric modelling of performance that provides principled representations of uncertainty to guide exploration. However, the $\O(N^3)$ scaling of \gls{gp} inference in the number $N$ of function evaluations has prevented the extension of these methods to many real problems of interest. Fitting complex models to real data often requires prohibitively large numbers of performance evaluations in order to effect the optimisation of model parameters. We make two central contributions towards these challenges.

\paragraph{Scalable \gls{bo}:}
Firstly, we propose a new \gls{bo} algorithm that makes use of advances in sparse \glspl{gp} to deliver more benign scaling. This enables the tackling of challenging algorithm configuration tasks and provides non-trivial acceleration on generic \gls{bo} problems. We provide explicit comparisons against the \gls{bo} state of the art for managing large numbers of evaluations,
\glspl{rfr} \cite{HutHooLey11,gramacy2013variable}. Specifically, we present evidence that \glspl{rfr} provide misleading uncertainty estimates that hinder exploratory optimisation relative to our algorithm.

\paragraph{Stochastic \gls{bo}:}
Our second contribution is introducing the use of \gls{bo} strategies for noisy optimisation to configure algorithms where performance evaluations are either uncertain or stochastic. This contribution is significant when considering the fitting of models to big data. The state of the art for such tasks relies upon evaluations of model fit on stochastically selected subsets of data, giving rise to algorithms including
stochastic optimisation \cite{Robbins1951},
stochastic gradient Langevin dynamics \cite{welling2011bayesian} and
stochastic variational inference \cite{hoffman2013stochastic}. Within our \gls{bo} framework, such evaluations on subsets are simply treated as evaluations of a latent performance curve corrupted by noise.
Unlike existing stochastic approaches, our \gls{bo} strategy uses these noisy objective evaluations to construct an explicit surrogate model even when \emph{gradients are unavailable} (as is often the case for black-box algorithms) or are excessively expensive. Coupled with our scalable models, we can take sufficient stochastic evaluations to explore and optimise for real, multi-modal, algorithm configuration problems.

\section{Bayesian Optimisation}
\label{sec:bayesopt}

We frame our central problem, algorithm configuration, as one of \gls{go}. That is, we view the performance of an machine learning algorithm (such as its cross-validation error, or likelihood) as an expensive function of the algorithm's parameters, and apply an optimiser to this function in order to find the best settings of its parameters. \gls{go} is required to explore typically  non-convex, complex parameter spaces using a parsimonious number of function evaluations.

Bayesian optimisation \cite{brochu2010tutorial,jones2001taxonomy} applies probabilistic modelling to global optimisation. Explicitly, we define the \gls{go} task as finding the minimum of an objective function $f(x)$ on some bounded set $\sX \subset \reals^d$, given only noise-corrupted observations of the objective, $y(x) = f(x) + \epsilon$, where $\epsilon \sim \N(0, \sigma^2)$. We consider the setting in which the expense of evaluating $y(x)$, and the unavailability of gradients of $y$, motivate the use of sophisticated means of iteratively selecting function evaluations. Specifically, we use a probabilistic model $p(f, y)$ as a surrogate for $f$, allowing evaluations from across the domain $\sX$, and the uncertainty inherent in $y$, to inform future selections. We define the set of information available after the $n$th function evaluation as $\D_n \coloneqq \bigl\{x_i,y(x_i) \mid i = 1, \ldots, n \bigr\}$.

Coupled with the probabilistic surrogate is an acquisition function $\lambda(x\mid \D)$ (often interpretable as an expected loss function), used as a means of choosing each successive function evaluation, as
$
    x_{i+1}
    =
    \argmax_{\xa} \lambda(\xa\mid \D)
$.
While this introduces a new optimisation problem, acquisition functions are chosen to admit observations of the gradient and Hessian, and to be trivially cheap to evaluate relative to the cost of the objective itself. There are a wide range of options for the choice of acquisition function \cite{brochu2010tutorial}. In this work, given our goal of managing noisy likelihood evaluations, we use the noise-tolerant expected improvement acquisition function of \cite{osborne2009gaussian}. Specifically, defining $\ya = y(\xa)$, ${f}_{\ast} = f({x}_{\ast})$ and $\nu$ as an appropriately small threshold, we choose
\begin{equation}
    \lambda(\xa\mid \D)
    \coloneqq
    \eta \int_{\eta}^{\infty} p(\ya \mid \D)\, \mathrm{d}\ya
    +
    \int_{\infty}^{\eta} \ya\, p(\ya \mid \D)\, \mathrm{d}\ya
    \text{;\quad}
    \eta \coloneqq \min_{{x}_{\ast} :
                            \mathbb{V}\bigl[ p({f}_{\ast} \mid \D) \bigr]
                            <
                            \nu^2
                        }
        \mathbb{E}\bigl[ p({f}_{\ast} \mid \D) \bigr].
\label{eq:acq_fn}
\end{equation}
That is, our acquisition function is the expected lowest function value about which we are sufficiently confident in (with confidence specified by $\nu$) after evaluating at $\xa$. Below, we will describe our extension of \gls{bo} to robustly manage many function evaluations, and to perform stochastic optimisation.

\section{Sparse Regression}

\subsection{Sparse Models}

The \gls{gp} is widely used in regression, classification and other machine learning algorithms  as a prior over functions due to its conceptual simplicity and full Bayesian treatment of uncertainty, providing not just a prediction of a value but also an associated estimate of the uncertainty. A \gls{gp}'s behaviour is defined by its covariance function $k(\mathbf{x}, \mathbf{x'})$. The function $k$ for $x, x' \in \sX$ maps $\sX \times \sX \mapsto \mathbb{R}$, and is known as a kernel. For two input or feature matrices $\mathbf{X}\in \mathbb{R}^{n \times d}$ and  $\mathbf{X}'\in \mathbb{R}^{m \times d}$ the $n \times m$ covariance matrix $K(\mathbf{X}, \mathbf{X}')$ is the Gram matrix made by the pairwise mapping of $k(\cdot, \cdot)$ to each row in $\mathbf{X}$ and $\mathbf{X}'$ (each row corresponding to a feature vector). For Gaussian noise (with variance $\sigma^2$) corrupted values $\mathbf{y}$ observed at $\mathbf{X}$ from a function $f \sim \mathcal{GP}(\mathbf{0}, K)$, we can express the distribution over $\mathbf{y}$ as
$
p(\mathbf{y} \mid \mathbf{X})
=
\mathcal{N}(y \mid \mathbf{0},\ \mathbf{K} + \mathbf{I}\sigma^2)
$.
Predictions $\mathbf{f}_*$ at inputs $\mathbf{X}_*$ using the \gls{gp} are
available in closed form \cite[Appendix B]{Bishop}.
The elegance of the mathematics underlying the prediction and marginalisation of the \gls{gp} comes at computational cost. Evaluation of the likelihood of the model scales as $\mathcal{O}(n^3)$ for $n$ samples, and prediction scales as $\mathcal{O}(pn^2)$ for $p$ predictions.

There is much prior art on improving the scaling of a \gls{gp} with $n$. Most are based on reducing the rank of the covariance matrix, rendering the matrix inversion $\mathcal{O}(m^3)$, where $m$ is the new rank.

\paragraph{FITC:} Perhaps the most widely used method is the \gls{fitc} method  \cite{Naish-Guzman2007}. This method uses an inducing matrix
$\bar{\mathbf{X}}$ (whose rows are feature vectors describing the locations of \emph{inducing points}) with associated latent values $\mathbf{u}$ to restrict the bandwidth of the kernel, forcing information exchange between the training and test data to pass through these points rather than the infinite bandwidth link of a full \gls{gp}. In the implementation used in this paper, we select a set of inducing points on a linear grid, optimise the \gls{gp} hyper-parameters, then jointly optimise the hyper-parameters and the inducing inputs. We found this to give superior results to performing the joint optimisation initially, and to be more accurate than simply fixing the inducing points.

\paragraph{Nystr\"{o}m-\acro{GP}:} A similar method to \gls{fitc} is the Nystr\"{o}m approximation \cite{Reece2013}. A kernel $k$ can be expressed by an infinite weighted sum of orthonormal bases
$
k(\mathbf{x}, \mathbf{x}') = \sum_{j=1}^\infty \mu_j^\phi \phi_j(\mathbf{x}) \phi_j(\mathbf{x}')
$
where $\phi_j$ and $\mu_j$ are the $j$th eigenfunctions and eigenvalues (henceforth `eigenpairs' when considered together) of the kernel under a distribution $p$ such that
$
\int k(\mathbf{x}, \mathbf{x}')\phi_j(\mathbf{x}')p(\mathbf{x}')d\mathbf{x}' = \mu_j^\phi \phi_j(\mathbf{x}).
$
It is rarely feasible to evaluate an infinite sum, so we consider only the $m$ most significant eigenpairs, which has the effect of reducing the frequency response of the kernel. Smoother kernels such as the \gls{se} require fewer components than rougher ones such as the Mat\'{e}rn \sfrac{3}{2}.

Following the recommendations in   \cite{Reece2013}, in the implementation used in this paper we maintain an area of interest $\mathcal{S}$ in which we wish to approximate the kernel and a representative sample set $\mathbf{S} \in \mathcal{S}$ of cardinality $L$. For a stationary kernel as used in this paper uniform sampling is acceptable, however for non-stationary kernels denser sampling in regions with shorter lengthscales is advised. We use a uniform probability density over $\mathcal{S}$ because we have no prior knowledge of the importance of different regions. We construct the Gram matrix by pairwise evaluation of the full kernel $k$ on the elements of $\mathbf{S}$, and find the eigenpairs ($\mu$ and $\mathbf{v}$). The eigenpairs of the Gram are used to construct the approximate eigenfunctions
$
\tilde{\phi_j}(\mathbf{x}) = \sfrac{\sqrt{L}}{\mu_j} K(\mathbf{x}, \mathbf{S}) \mathbf{v}_j
$
and values $\mu_j^\phi = \sfrac{\mu_j}{L}$. Using these approximate eigenpairs we can express the truncated approximation as
$
k(\mathbf{x}, \mathbf{x}') \approx \sum_{j=1}^m \mu_j^\phi \tilde{\phi_j}(\mathbf{x}) \tilde{\phi_j}(\mathbf{x}').
$
We refer to this approximation as the \emph{Nystr\"{o}m-\acro{GP}}.

\paragraph{Laplacian-\acro{GP}:} A similar result to \cite{Reece2013} is arrvied at in \cite{Solin2014} by finding the eigenfunctions of the Laplacian within a specific domain. They define a covariance operator
$
\mathcal{K}\phi = \int k(\cdot, \mathbf{x}')\phi(\mathbf{x}')\ d\mathbf{x}'
$
and, for a isotropic function $k(\mathbf{x}, \mathbf{x}') \coloneqq k(\|\mathbf{r}\|)$ in a compact set $\Omega \subset \mathbb{R}^d$, expand the operator $\mathcal{K}$ into a series of Laplacian operators. They use this to generate a series expansion of the kernel $k$, leading (by analogous argument to that used in the Nystr\"{o}m \gls{gp}) to the truncated approximation
$
k(\mathbf{x}, \mathbf{x}') \approx \sum_{j=1}^m S(\sqrt{\lambda_j}) \phi_j(\mathbf{x})\phi_j(\mathbf{x}'),
$
where $\lambda_j$ and $\phi_j$ are the eigenpairs of the Laplacian in a given domain and $S(\cdot)$ is the spectral density of the covariance function. In general, the eigenpairs of the Laplacian have simple closed form solutions. For example, in one dimensional Cartesian coordinates they correspond to the Fourier basis. In order to find the eigenpairs of the Laplacian, boundary conditions must be specified. Dirichlet boundary conditions force the response to zero at the edges of $\Omega$, leading to some distortion near the edges of the space. Solin et al. recommend defining an area of interest and setting the boundary conditions slightly beyond this. In this paper we refer to this approximation as the \emph{Laplacian-\acro{GP}}.

Both the Nystr\"{o}m-\acro{GP} and the Laplacian-\acro{GP} lead to an rank $m$ approximation to the full covariance $\mathbf{K} \approx \Phi^T \Lambda \Phi$, allowing inversion in $\mathcal{O}(m^3)$ using the matrix inversion lemma.

\paragraph{Sparse spectrum \acro{gp}:}   \citeauthor{Lazaro-Gredilla2010}   \cite{Lazaro-Gredilla2010} provide an alternative method to select the eigenpairs by Monte-Carlo sampling of the kernel function's spectral density. This method is found to compare poorly (for equivalent covariance rank $m$) to the methods presented in both   \cite{Reece2013} and   \cite{Solin2014} due to the random selection of spectral points.

\paragraph{Random forest regressors:}

\glspl{rfr} are recommended as an alternative to the \gls{gp} for \gls{bo} by Hutter et al.  \cite{HutHooLey11}. They note the computational load of the \gls{gp}, and the possibly sub-optimal management of changepoints. We found that the latter criticism is of limited importance for the real objective functions we consider,
and the sparse \glspl{gp} considered in this paper aims to rectify the former. In this paper, we use the \gls{rfr} from scikit-learn  \cite{Pedregosa2011} with 30 trees. The posterior mean and variance were taken as the mean and variance of the outputs of the regressors.

\subsection{Model Accuracy and Computational Complexity}\label{sec:posterior}

\begin{figure}[ht]
    \begin{minipage}[b]{0.45\linewidth}
        \centering
        \includegraphics[width=\textwidth]{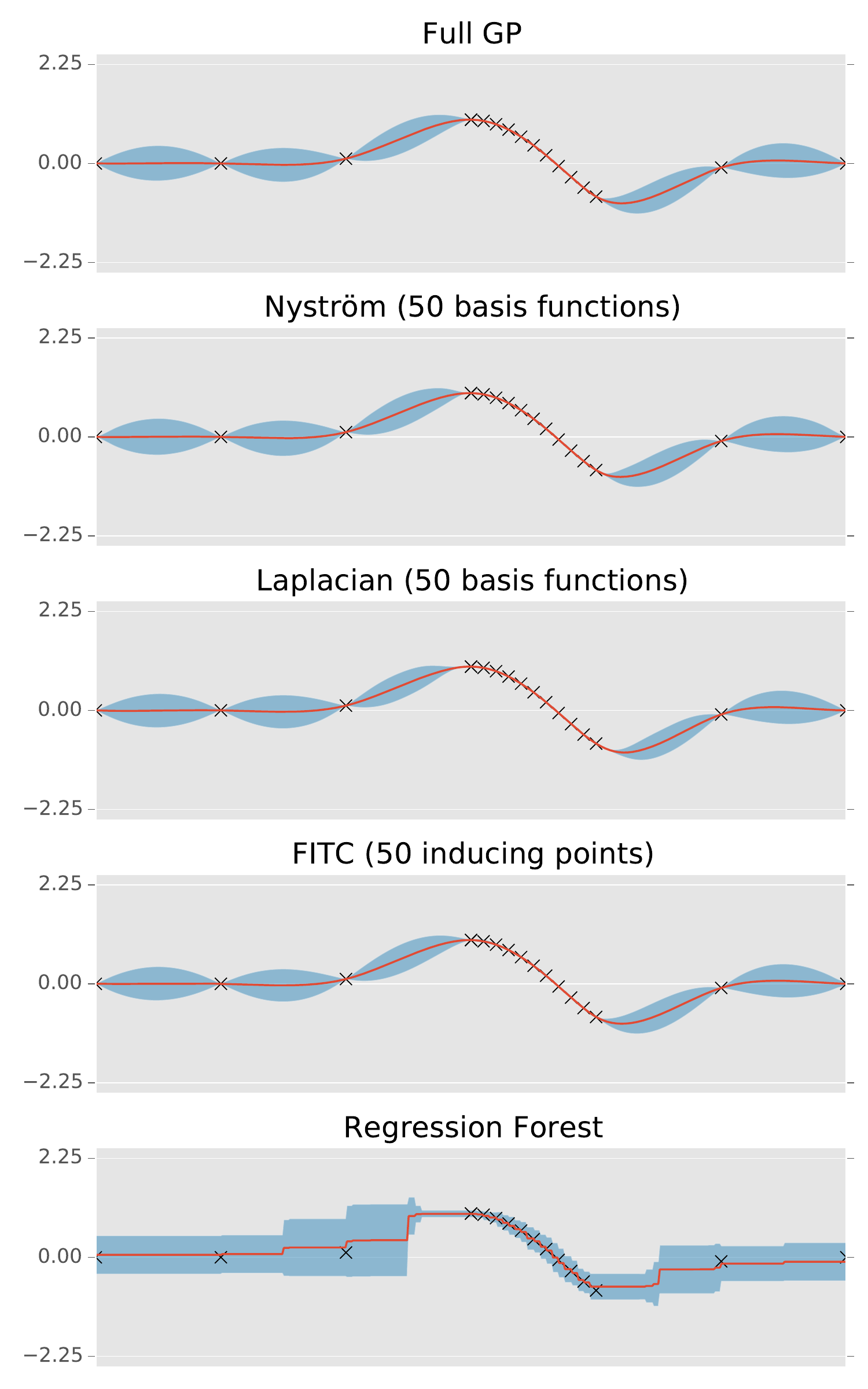}
        \caption{Mean and variance of the GP, spectral GPs and other methods}
        \label{fig:posteriors}
    \end{minipage}
    \hspace{0.5cm}
    \begin{minipage}[b]{0.45\linewidth}
        \centering
        \includegraphics[width=\textwidth]{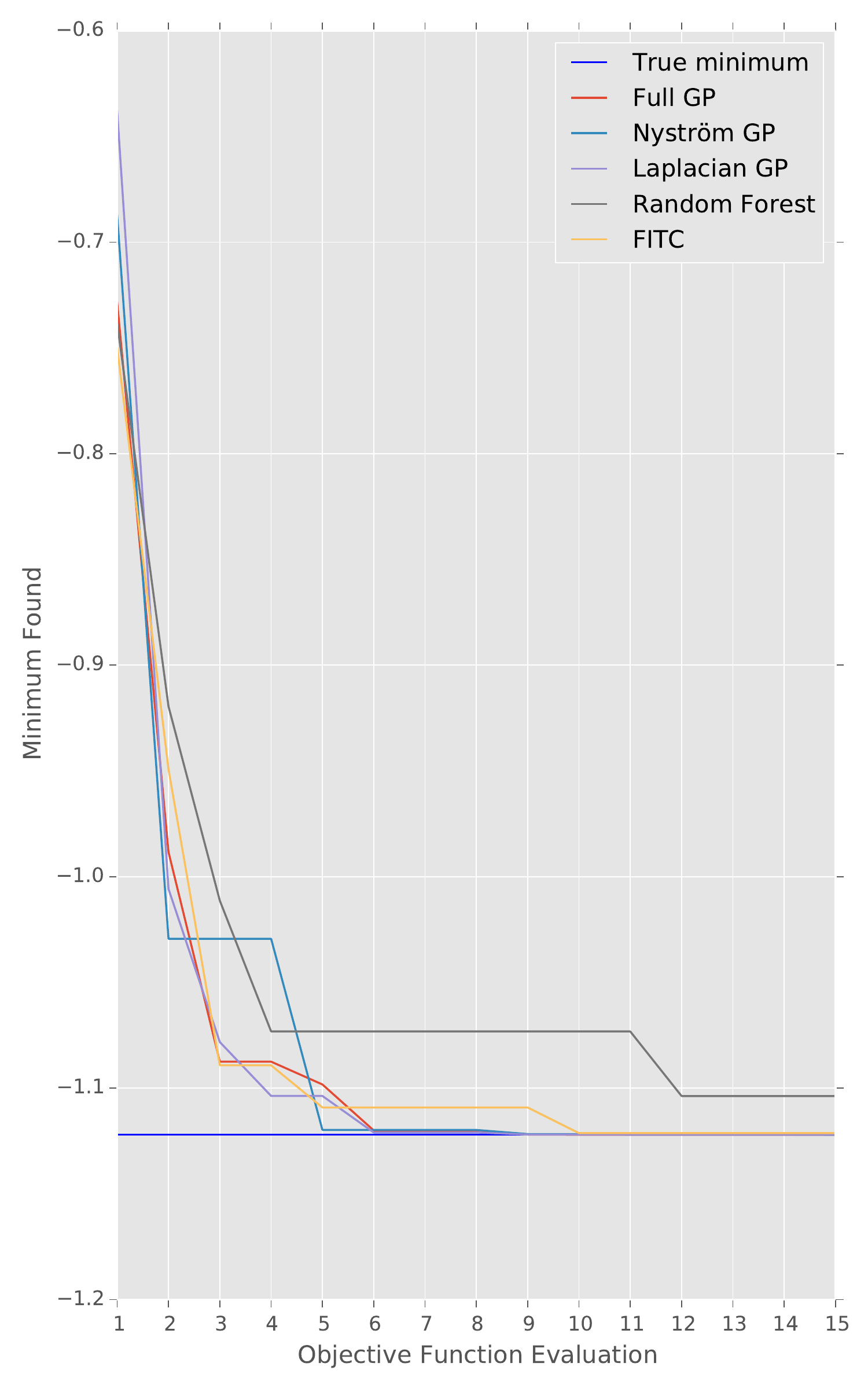}
        \caption{Convergence of different regressors to the true minimum.}
        \label{fig:simple_bo_converge}
    \end{minipage}
\end{figure}

Figure \ref{fig:posteriors} shows the posteriors of the Full \gls{gp} and the Nystr\"{o}m, \gls{fitc} and Laplacian approximations discussed above, along with a \gls{rfr}. There is very little to distinguish the \gls{gp} approximations given a suitable number of inducing points or eigenpairs. The regression forest shows reasonable behaviour, by reducing its variance in densely sampled regions or regions where the mean is flat, and increasing it in sparsely sampled or steep regions. This dependence on gradient may be problematic because sparsely sampled flat regions will have a spuriously low variance. The effects of the posterior inaccuracy on \gls{bo} are quantitatively evaluated in Section \ref{sec:bo_posterior}.

The computational complexities of the 5 methods considered here are listed in Table \ref{tab:complexity}. A point of note is that the \emph{Laplacian approximation scales poorly with dimensionality} because it requires a dense spectrum. If $m$ is the cardinality of the basis needed for a good representation in one dimension, for a $d$ dimensional space one would need $m^d$ bases. This presents problems in high dimensions: fourteen basis functions in five dimensions would lead to a matrix with rank 537,824.

\begin{figure}[h]
        \centering
        \includegraphics[width=\textwidth]{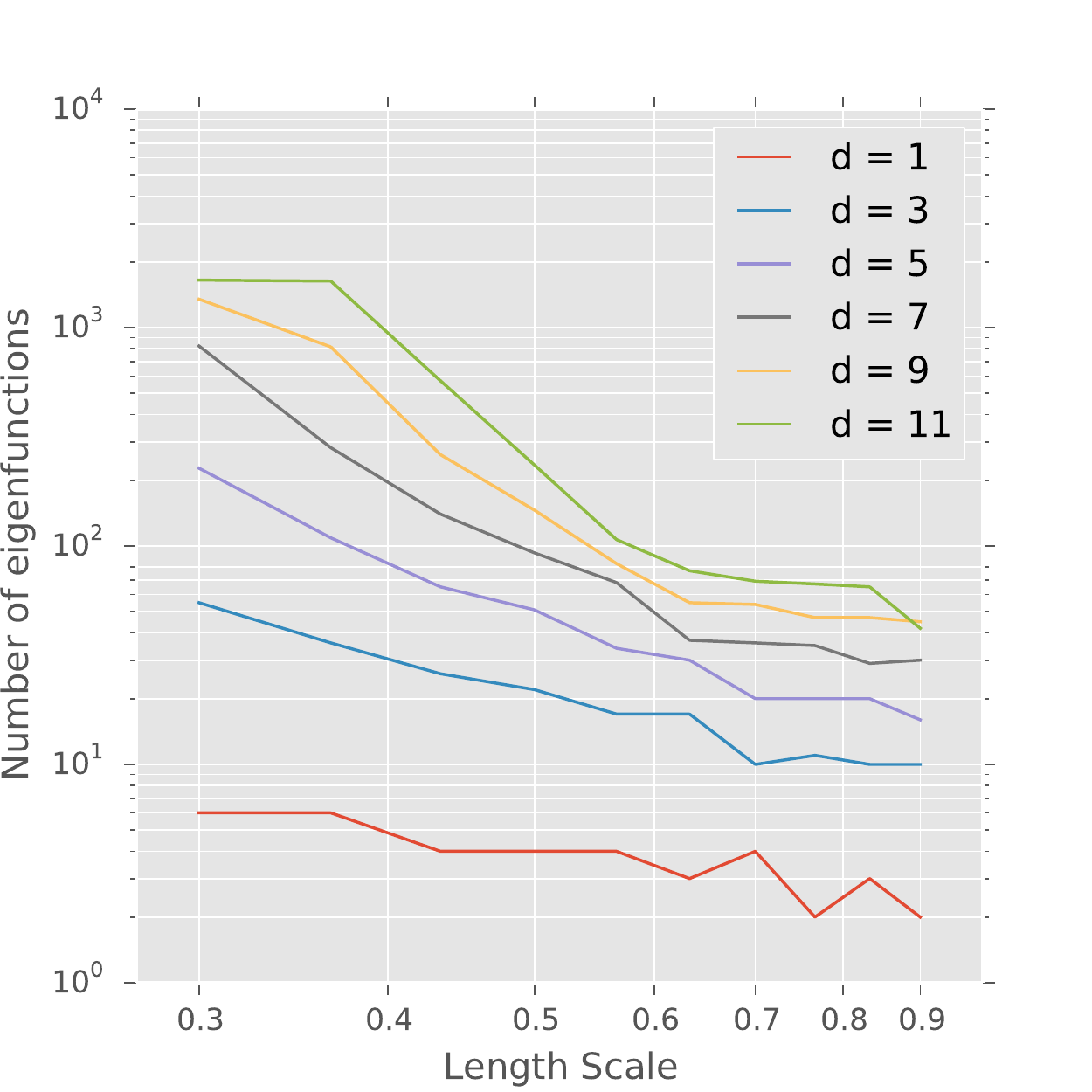}
        \caption{Scaling of number of Eigenfunctions with lengthscale and dimensionality.}
        \label{fig:eigensize}
 \end{figure}

The Nystr\"{o}m intelligently (if heuristically) selects the $m$ most important eigenpairs for a given space, and the \gls{fitc} approximation allows one to select the number of inducing points needed. In the case of the \gls{fitc} approximation, there is no clear way to select the optimal set of inducing points. The method of optimisation used in this paper is unreliable with many inducing inputs or in high dimension, and a na\"{i}ve grid of inducing points scales poorly with dimensionality. Incorrectly located inducing points can cause the posterior variance to become excessively large or small.
The Nystr\"{o}m method has more benign scaling with dimensionality.
 Figure \ref{fig:eigensize} shows the relationship between the number of eigenpairs and the lengthscale of the \gls{gp} kernel. These values were generated in a space $[0, 1]^d$, with length scales between $1$ and $0.1$. For very high dimensional spaces it would seem impractical to use a length-scale much below the size of the space.
It should be noted that we used the heuristic recommended in Reece et al.  \cite{Reece2013} and selected all eigenpairs with eigenvalues larger than $\sfrac{\text{max eig}}{100}$, where $\text{max eig}$ is the largest eigenvalue of the matrix.

We found that the logarithm of the number of basis functions needed scaled inversely with the logarithm of the length-scales of the \gls{gp}. In the worst case we tested (11 dimensions with length scale 0.3 in $[0, 1]^{11}$) we needed 1,500 bases to adequately represent the kernel. As length-scales become very short in relation to the input space, the exponential scaling begins to reduce the viability of the this model, however we have not found this to be a problem in practical algorithm configuration. Figure \ref{fig:eigensize} shows the log-log relationship between the lengthscale and the number of bases.
\begin{table}[h]
\begin{center}
\begin{tabular}{lll}
\toprule
Regressor & Prediction complexity & Learning complexity\\
\midrule
\gls{gp} & Initial $\mathcal{O}(n^3)$ then $\mathcal{O}(pn^2)$ & $\mathcal{O}(hn^3)$ \\
Nystr\"{o}m & Initial $\mathcal{O}(m^3)$ then $\mathcal{O}(pm^2)$ & $\mathcal{O}(hs^3m^3)$\\
Laplacian & Initial $\mathcal{O}(m^3)$ then $\mathcal{O}(pm^2)$ & Initial $\mathcal{O}(hnm^2)$ then $\mathcal{O}(hm^3)$\\
\gls{fitc} & $\mathcal{O}(pm^2)$ & $\mathcal{O}(nm^2)$\\
Regression Forest & $\mathcal{O}(ptc)$ & $\mathcal{O}\bigl(dtn\text{log}(n)\bigr)$\\

\bottomrule
\end{tabular}
\end{center}
\caption{Computational complexity of regressors. $n$ is the number of training points, $p$ is the number of prediction points, $s$ is the size of the sample set used to characterise the spectrum of the kernel, $h$ is the number of training steps when optimising the hyper-parameters, $t$ is the number of trees making up the regression forest, $c$ is the number of decisions that must be made in a tree, $d$ is the dimensionality of the input space and $m$ is the rank of the low-rank covariance matrix. \gls{gp} results from  \cite{Rasmussen2005}, Nystr\"{o}m from   \cite{Reece2013}, Laplacian from   \cite{Solin2014}, \gls{fitc} from   \cite{Naish-Guzman2007} and \gls{rfr} from  \cite{Breiman1984}.}
\label{tab:complexity}
\end{table}


\section{Sparse Bayesian Optimisation on a Mixture of Gaussians}
\label{sec:bo_posterior}


We tested each of the five regressors above as the response surface within \gls{bo} to optimise a toy mixture of Gaussians with 5 local minima and one global minimum in [-5, 5]. For the first test, we used two randomly sampled points to `prime' the regressors. The full \gls{gp}, Laplacian, \gls{fitc} and Nystr\"{o}m approximations all converged to the minimum within 10 iterations. The \gls{rfr} approximation did not perform well. The use of the empirical mean and variance in the \gls{rfr} caused a large amount of the function space to be given zero variance and so was effectively ignored as the regressor throughly explored a local minima. As a further experiment, we primed the regressors with 200 randomly sampled points in [-5, 5]. With this density of samples all methods converged to the global minimum, however again the \gls{rfr} took considerably longer.

The posterior of the \gls{gp} (and its various sparse approximations) allows the regressor to be bootstrapped from a minimal starting set of sample points. The initial uncertainty strongly promotes exploration. This is in contrast to the \gls{rfr}, which can be excessively certain in sparsely sampled or flat regions. Incorrect estimates of the hyper-parameters can cause poor fitting of the function in all of the \gls{gp} models, causing exploration or exploitation to proceed incorrectly. The \gls{fitc} approximation is most affected by this due to the additional optimisation required by the inducing inputs which may cause it to under- or over-estimate the variance, as shown by its slower convergence in Figure \ref{fig:simple_bo_converge}.

We have found little difference in the performance between the \gls{fitc}, Laplacian and Nystr\"{o}m approximations (the Laplacian model is slightly more efficient in one dimension, while the \gls{fitc} is more prone to incorrectly estimating the variance if the inducing inputs are incorrectly located, causing the delayed convergence shown in Figure \ref{fig:simple_bo_converge}). We will concentrate on the Nystr\"{o}m approximation henceforth, due to its better scaling with dimensionality and more reliable variance.

%
%

\section{Stochastic Bayesian Optimisation}

Stochastic inference is a technique where the likelihood is evaluated on subsets of the data. Methods in this family include
stochastic optimisation \cite{Robbins1951},
stochastic gradient Langevin dynamics \cite{welling2011bayesian} and
stochastic variational inference \cite{hoffman2013stochastic,hensman2013gaussian}. These methods inspire \emph{Stochastic Bayesian Optimisation,} henceforth known as \gls{stoat}. Here, we make noisy observations of the likelihood of a large \gls{ml} model by evaluating it on subsets of the data. We use the probabilistic power of the \gls{gp} and the large-data capabilities of the Nystr\"{o}m approximation to make many observations of the likelihood with different subsets of the data at each step of the \gls{bo} algorithm.
Our approach provides a means of global exploration using stochastic likelihood evaluations that complements the local, gradient-driven, optimisation enabled by existing stochastic approaches. Unlike these approaches, we do not make use gradient observations, allowing us to consider real, black-box algorithm configurations for which gradients are unavailable or excessively expensive. Our approach also permits the global exploration of complex likelihood surfaces, reducing the risk  presented by local minima.

\subsection{Optimisation of the Branin function}

We tested the performance of both traditional \gls{bo} and Nystr\"{o}m \gls{bo} in optimising the Branin function \cite{Dixon1978}, a standard function for testing machine learning algorithms. With $200$ basis functions, the performance of the full \gls{gp} and Nystr\"{o}m approximations were indistinguishable. By wall clock time, the full \gls{gp} performed better initially (when the time taken to compute the Cholesky decomposition was less than the eigendecomposition of the Gram matrix), however the Nystr\"{o}m method showed better scaling as $n$ increased. We also found that the Nystr\"{o}m method was more robust to conditioning errors caused by making multiple local samples.

 \emph{To test the hypothesis that we can perform \gls{go} with noisy observations,} we re-ran the experiment with the evaluations of the Branin function corrupted by Gaussian noise with $\sigma^2 = 5$, which is very large compared to the range of function values around the Branin's three minima.
 Each experiment was limited to 1000 seconds of wall-clock time.
 At each step we made 50 observations of the noisy Branin function. These were either passed to the \gls{gp} directly (`full' mode), or averaged and passed to the \gls{gp} as a single less noisy observation (`average' mode). With 50 noisy evaluations per \gls{bo} evaluation, and 20-30 seconds per evaluation, the `full' mode gathered between 1,500 and 2,500 samples, while the `average' mode gathered 30 to 50. In tests, we found the full mode to outperform the average mode with little computational overhead.
In additional, we drew 400 points from a Sobol sequence, evaluated the noisy Branin on these and passed them to the \gls{gp} as a pre-sample, to allow it to concentrate more closely on exploring low regions of the space and learn hyper-parameters from a larger initial set. In the most extreme case \gls{stoat} efficiently performed \gls{bo} with nearly 4,000 data points.

For each of the algorithms, we started a local minimiser at each of the three true minima of the Branin function and let them run to convergence, finding the local minima of the response surface nearest to each of the Branin minima (in all cases one of these three points was also the global minimum of the response surface). We take this global minimum of the response surface as our estimate of $\nu$, as discussed in Section \ref{sec:bayesopt}.
We used the `Gap' measure of performance to compare the methods \cite{Huang2006}:
\begin{equation}
G \coloneqq
\frac{y\left(x^\text{init}\right) - y\left(x^\text{best}\right)}
{y\left(x^\text{init}\right) - y\left(x^\text{opt}\right)}
\label{eq:gap}
\end{equation}.
Table \ref{tab:branin} shows the maximum gap, mean gap, and the best minimum found by \gls{stoat} and the standard \gls{bo} algorithm. We also tested \gls{cmaes} \cite{Hansen1996} and \gls{direct} \cite{Jones1993} however found that these did not converge on this very noisy objective.

\begin{table}[tbh]
\begin{center}
\begin{tabular}{lrrr}
\toprule
Method & Max Gap & Mean Gap & Best minimum found \textit{(true minima)}\\
\midrule
\gls{bo} & 0.926 & 0.831 & 2.15 \textit{(0.398)}\\
Stochastic \gls{bo} & \textbf{0.997} & \textbf{0.965} &  \textbf{0.472} \textit{(0.398)}\\
\bottomrule
\end{tabular}

\end{center}

\caption{Performance of the algorithms minimising the Branin function with observation noise.}
\label{tab:branin}
\end{table}

\section{Stochastic Bayesian Configuration of Model Parameters on Energy Data}

AgentSwitch \cite{AS2013} is a project that aims to assist people in selecting the most economical energy tariff for their expected electricity use. A \gls{gp} model is used to predict the energy that will be used by a household. The posterior prediction of this \gls{gp} is  used to inform group bidding for energy tariffs, to ensure that a user pays the lowest price for their power. The posterior variance is particularly useful here, because it allows the group to understand the risk of going for a cheaper, fixed use contract compared to a more expensive flexible plan. To replicate the work in \cite{AS2013} we used on household power use data\footnote{http://archive.ics.uci.edu/ml/datasets/Individual+household+electric+power+consumption}. \emph{We will use our algorithm to fit the parameters of this complex model to big data.  }


An inspection of Fourier transform of the power use data may allow one to find the major frequencies. There is a clear peak between 350 and 500 days (the resolution prohibits more accuracy) and a multitude of peaks at higher frequencies. It is not clear which is the best to select. In addition, the data is not equally spread in time, with some elements missing, making any Fourier transform an approximation of the true spectrum.

The authors of \cite{AS2013} fix the periodicity hyper-parameter {a-priori}
to one day and fit the other parameters using \gls{mle2} on subsets of the data. From inspection of the log-likelihood surface (and the spectrum computed above), we can see that there is a strong periodic component around one year, however the lower period minima is highly multi-modal. To adequately explore this surface would require multiple restart at different initial hyper-parameter values, which quickly approaches the computational burden of true \gls{bo}. This authors of \cite{AS2013} chose not to optimise their period with \gls{mle2} for this reason and set it to a fixed number; \gls{stoat} automatically balances exploration and exploitation, and can learn the two periods on large, real data in comparable time to a multi-start optimisation of \gls{mle2} \emph{learning only the non-periodic hyper-parameters}.

The dataset contains seven features describing the average of power use in the house in different ways, in addition to the date and time. The data was gathered at a resolution of one sample per minute. We down-sampled this to hourly averages, and selected just the `global active power' (power actually used throughout the household). In total we had 34,166 observations, with a single input dimension. We retained the last 5,000 entries for testing, and did not use these for learning the hyper-parameters. Our training set consisted of 29,166 samples. Computing the full likelihood on this data is extremely impractical on reasonable computing hardware, requiring 50 gigabytes of RAM simply to \emph{store} the Gram matrix.

\begin{figure}[t]
\begin{center}
    \begin{subfigure}[t]{0.45\textwidth}
        \centering
        \includegraphics[width=\textwidth]{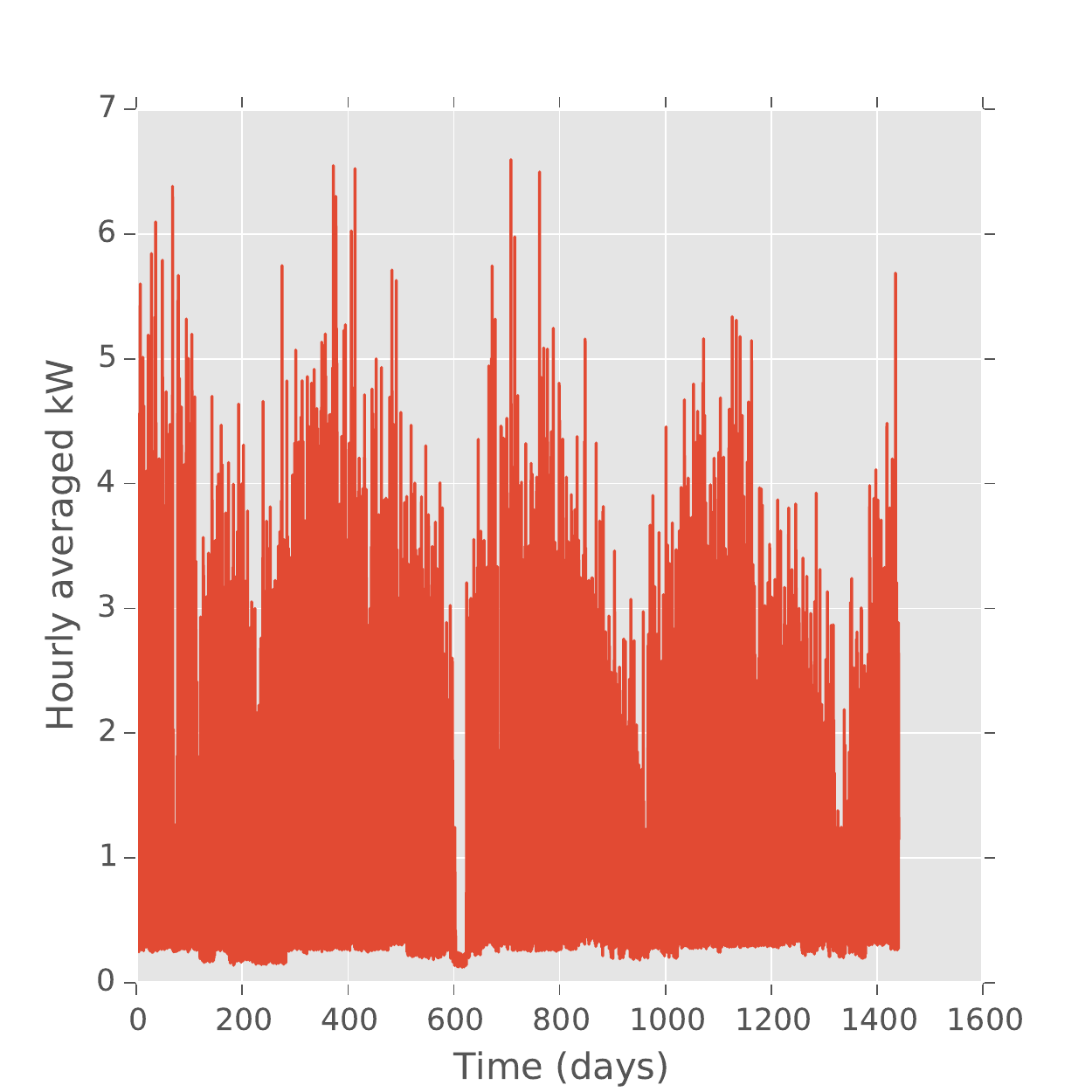}
        \caption{47 months of power usage, showing yearly cycle.}
        \label{fig:real_data}
    \end{subfigure}
    \hspace{0.5cm}
    \begin{subfigure}[t]{0.45\textwidth}
        \centering
        \includegraphics[width=\textwidth]{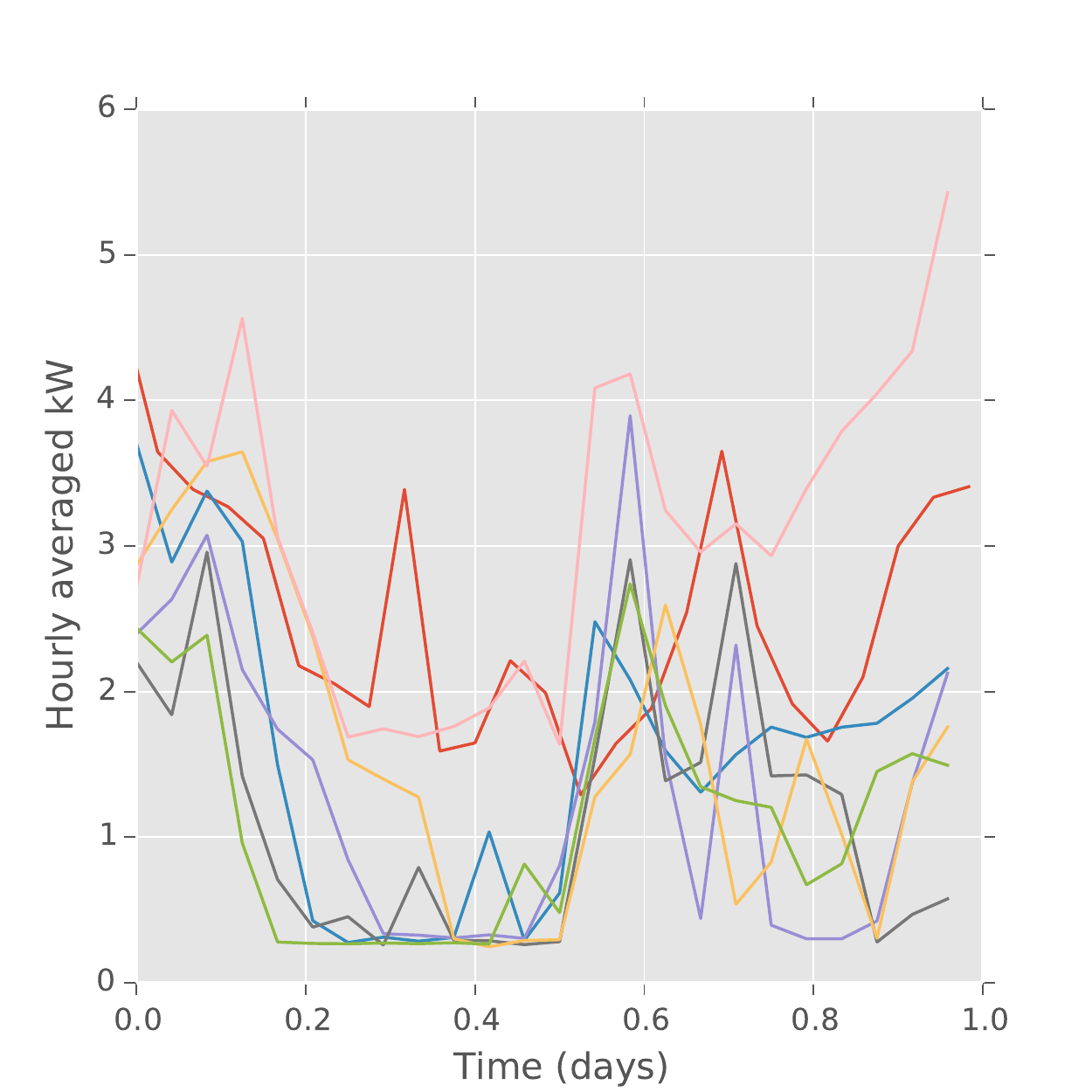}
        \caption{Seven single-day power traces, overlaid.}
        \label{fig:real_data_day}
    \end{subfigure}
    \caption{Hourly averaged household power use.}
    \end{center}
\end{figure}

We used \gls{stoat} to learn a the periods of a sum of periodic \gls{se} kernels on this data. We constrained the periods to lie within $[0.1, 10] \times [10, 1000]$. Our noisy likelihood observations were evaluated on 1,000 samples from the training data. At each step we made 10 of these stochastic observations (each on a different random subset of the training data without replacement). We ran the experiments on a laptop computer with a 2.3 GHz Intel i7 CPU. In addition to the 10 stochastic samples at each iteration, before beginning our \gls{bo} sequence we generated 600 space filling points from a Sobol sequence and evaluated the noisy likelihood at each of these.

The pre-sampling allowed the \gls{gp} to quickly become certain about the hyper-parameters, which reduced wasted exploration steps. The pre-sample step took around 5 minutes, however once it was complete the larger period quickly converged to 382 days, approximately one year. The shorter period oscillated between 1.3 and 1.5 days. After one hour, the optimiser had converged to 1.5. For comparison to \cite{AS2013}, we also optimised a single-periodic kernel using  \gls{stoat}. After the pre-sample, this converged after a few steps to a period of 378 days. Marginalisation of the hyper-parameters as recommended in \cite{osborne2009gaussian} (\gls{bq}) or \cite{snoek-etal-2012b} (\gls{mcmc}) may reduce the need to pre-sample, at the expense of additional computational load. Each iteration of the \gls{bo} algorithm took between 20 and 30 seconds to complete. Including the pre-sample, the number of measurements made was between 1,500 and 2,500 with no noticeable slow down as $n$ grew.

\begin{table}[h]
\begin{center}
\begin{tabular}{lr}
\toprule
Method & Test data loglikelihood\\
\midrule
\gls{stoat} learned double periodic & \textbf{-7.25}\\
\gls{stoat} learned single periodic & -7.39\\
AgentSwitch {a-priori} single periodic & -7.40\\
Aperiodic \gls{gp} &  -9.22\\

\bottomrule
\end{tabular}
\end{center}

\caption{Data log-likelihood on real electricity use data of models learned using our method, {a-priori} setting of periods and na\"{i}ve a-periodic \gls{gp}.}
\label{tab:money_shot}
\end{table}
To test our results, we compared the predictive log-likelihoods on held out test-data of our dual periodic and single periodic kernels, \cite{AS2013}'s single periodic kernel and a simple, aperiodic \gls{se} kernel. The results in Table \ref{tab:money_shot} show the model using our parameters tuned by our algorithm outperforming both \cite{AS2013} and the aperiodic \gls{gp}. Additionally, our algorithm tuner's ability to quickly find the second period at 1.5 days substantially improves the predictive performance when compared to simpler models, even when searching the highly multi-modal first dimension.

%

\section{Conclusion}

Using \gls{stoat} on a consumer grade laptop, we have quickly optimised the parameters of an \gls{ml} algorithm of such computational complexity that \emph{we cannot evaluate the likelihood on the full data}. On real, noisy data our algorithm quickly converges to the large global optimum in one dimension, and in two dimensions is able to find a second optimal location amongst many nearby local optima.

We extend the principled exploration of expensive functions developed in the \gls{bo} and \gls{smbo} literature to allow noisy observations of an objective function. In real machine learning problems with computationally intractable likelihoods we are able to find the global optimum by evaluating the likelihood on subsets of the data, relying on the information handling properties of our sparse \gls{gp} to allow for this additional noise.


 \subsubsection*{Acknowledgments}
This work is supported by the UK Research Council (EPSRC) funded ORCHID Project EP/I011587/1.
Thanks to Chris Lloyd for his help with the \gls{fitc} approximation.
Thanks to the authors, packager and maintainers involved with the Julia language  \cite{Bezanson2012}.
\renewcommand*{\bibfont}{\small}
\printbibliography
\end{document}

%% file: glossary.tex
\newacronym
[longplural={Marginalised Gaussian processes}]
{mgp}{mgp}{Marginalised Gaussian process}
\newacronym
[longplural={Gaussian processes}]
{gp}{gp}{Gaussian process}
\newacronym{lda}{lda}{latent Dirichlet}
\newacronym{mab}{mab}{multi-armed bandit}
\newacronym{dpp}{dpp}{determinantal point process}
\newacronym{abc}{abc}{approximate Bayesian computation}
\newacronym{eq}{eq}{exponentiated quadratic}
\newacronym{se}{se}{squared exponential}
\newacronym{nigp}{nigp}{noisy input Gaussian process}
\newacronym{ard}{ard}{automatic relevance detection}
\newacronym{iid}{i.i.d.}{independently identically distributed}
\newacronym{ibcc}{ibcc}{independent Bayesian classifier combination}
\newacronym{mle3}{mle-iii}{type 3 maximum likelihood}
\newacronym{mle}{mle}{maximum likelihood estimation}
\newacronym{mle2}{mle-ii}{type 2 maximum likelihood estimation}
\newacronym{mcmc}{mcmc}{Markov chain Monte-Carlo}
\newacronym{skld}{skld}{symmetric Kullback-Leibler divergence}
\newacronym{gpgo}{gpgo}{Gaussian process global optimisation}
\newacronym{awgn}{awgn}{additive white Gaussian noise}
\newacronym{bq}{bq}{Bayesian Quadrature}
\newacronym{nn}{nn}{neural network}
\newacronym{rq}{rq}{rational quadratic}
\newacronym{gps}{gps}{global positioning system}
\newacronym{mlhgp}{mlhgp}{most likely heteroscedastic Gaussian process}
\newacronym{glm}{glm}{generalised linear model}
\newacronym{gpml}{gpml}{Gaussian process machine}
\newacronym{vb}{vb}{variational Bayes}
\newacronym{psd}{psd}{positive semidefinite}
\newacronym{adc}{adc}{analogue to digital converter}
\newacronym{dac}{dac}{digital to analogue converter}
\newacronym{co}{co}{carbon monoxide}
\newacronym{ppm}{ppm}{parts per million}
\newacronym{icml}{icml}{international conference on machine learning}
\newacronym{bo}{bo}{Bayesian optimisation}
\newacronym{uav}{uav}{unmanned aerial vehicle}
\newacronym{hd}{hd}{high definition}
\newacronym{ei}{ei}{expected improvement}
\newacronym{pi}{pi}{probability of improvement}
\newacronym{ucb}{ucb}{upper confidence bound}
\newacronym{sdpp}{s-dpp}{structure determinantal point process}
\newacronym{soc}{soc}{state of charge}
\newacronym{acfr}{acfr}{Australian centre for field robotics}
\newacronym{uart}{uart}{universal asynchronous receiver/transmitter}
\newacronym{ndk}{ndk}{native development kit}
\newacronym{bbo}{bbo}{batch Bayesian optimisation}
\newacronym{fitc}{fitc}{fully independent training conditional}
\newacronym{direct}{direct}{dividing rectangles}
\newacronym{el}{el}{expected loss}
\newacronym{rfr}{rfr}{random forest regressor}
\newacronym{nll}{nll}{negative log-likelihood}
\newacronym{ml}{ml}{machine learning}
\newacronym{go}{go}{global optimisation}
\newacronym{cmaes}{cmaes}{covariance matrix adaptive evolution strategy}
\newacronym{stoat}{stoat}{STOchastic Algorithm Tuning}
\newacronym{smbo}{smbo}{Sequential Model Based Optimisation}